\documentclass[10pt,twocolumn,letterpaper]{article}

\usepackage{cvpr}              

\usepackage{graphicx}
\usepackage{amsmath}
\usepackage{amssymb}
\usepackage{booktabs}
\usepackage{comment}

\usepackage{cite}
\usepackage{amsfonts}
\usepackage{graphicx}
\usepackage{textcomp}
\usepackage{xcolor, array, colortbl}
\usepackage[utf8]{inputenc}
\usepackage[T1]{fontenc}      
\usepackage{url}                    
\usepackage{nicefrac}    
\usepackage{microtype}     
\usepackage{bbding}
\usepackage{pifont}
\newcommand{\xmark}{\ding{55}} 
\newcommand{\cmark}{\ding{51}}

\usepackage{fancyhdr}
\usepackage{multirow}
\usepackage{rotating}
\usepackage{color}
\usepackage{bm}
\usepackage{enumerate} 
\usepackage{enumitem}
\usepackage{mathrsfs}
\usepackage{makeidx}           
\usepackage{multicol}  
\usepackage[subfigure]{tocloft}     
\usepackage{epsfig,latexsym}
\usepackage{psfrag}
\usepackage[ruled,vlined]{algorithm2e}
\usepackage{algorithmic}
\definecolor{lightgray}{rgb}{.91,.91,.91}
\definecolor{deepred}{rgb}{0.698,0.133,0.133}
\definecolor{blue}{rgb}{0,0,1}

\newcommand{\mf}{\mathbf}

\newcommand{\mr}{\mathrm}

\usepackage[pagebackref,breaklinks,colorlinks]{hyperref}

\newenvironment{myitemize}{\begin{list}{$\bullet$}
		{\setlength{\topsep}{1mm}
			\setlength{\itemsep}{0.25mm}
			\setlength{\parsep}{0.25mm}
			\setlength{\itemindent}{0mm}
			\setlength{\partopsep}{0mm}
			\setlength{\labelwidth}{15mm}
			\setlength{\leftmargin}{4mm}}}{\end{list}}

\usepackage[capitalize]{cleveref}
\crefname{section}{Sec.}{Secs.}
\Crefname{section}{Section}{Sections}
\Crefname{table}{Table}{Tables}
\crefname{table}{Tab.}{Tabs.}

\def\cvprPaperID{4286} 
\def\confName{CVPR}
\def\confYear{2023}

\begin{document}

\title{Federated Incremental Semantic Segmentation}

\author{Jiahua Dong\textsuperscript{1, 2, 3}\footnotemark[1]~, Duzhen Zhang\textsuperscript{4}\footnotemark[1]~, Yang Cong\textsuperscript{1, 2}\footnotemark[2]~, Wei Cong\textsuperscript{1, 2, 3}, Henghui Ding\textsuperscript{4}, Dengxin Dai\textsuperscript{4}\\
\textsuperscript{1}State Key Laboratory of Robotics, Shenyang Institute of Automation, \\ Chinese Academy of Sciences, Shenyang, 110016, China.\footnotemark[3]\\
\textsuperscript{2}Institutes for Robotics and Intelligent Manufacturing, \\ Chinese Academy of Sciences, Shenyang, 110169, China. \\
\textsuperscript{3}University of Chinese Academy of Sciences, Beijing, 100049, China. \\
\textsuperscript{4}ETH Z\"{u}rich, Z\"{u}rich, 8092,
Switzerland. \\
{\tt\small \{dongjiahua1995,\;congyang81,\;congwei45,\;henghui.ding\}@gmail.com, dai@vision.ee.ethz.ch }
}

\maketitle

\renewcommand{\thefootnote}{\fnsymbol{footnote}}
\footnotetext[1]{Equal contributions. 
\footnotemark[2]The corresponding author is Prof. Yang Cong.
$~~$\indent\footnotemark[3]This work was supported in part by the National Nature Science Foundation of China under Grant 62127807, 62225310 and 62133005.}

\begin{abstract}
Federated learning-based semantic segmentation (FSS) has drawn widespread attention via decentralized training on local clients. However, most FSS models assume categories are fixed in advance, thus heavily undergoing forgetting on old categories in practical applications where local clients receive new categories incrementally while have no memory storage to access old classes. Moreover, new clients collecting novel classes may join in the global training of FSS, which further exacerbates catastrophic forgetting. To surmount the above challenges, we propose a {\textbf{F}}orgetting-{\textbf{B}}alanced {\textbf{L}}earning (\textbf{FBL}) model to address heterogeneous forgetting on old classes from both intra-client and inter-client aspects. Specifically, under the guidance of pseudo labels generated via adaptive class-balanced pseudo labeling, we develop a forgetting-balanced semantic compensation loss and a forgetting-balanced relation consistency loss to rectify intra-client heterogeneous forgetting of old categories with background shift. It performs balanced gradient propagation and relation consistency distillation within local clients. Moreover, to tackle heterogeneous forgetting from inter-client aspect, we propose a task transition monitor. It can identify new classes under privacy protection and store the latest old global model for relation distillation. Qualitative experiments reveal large improvement of our model against comparison methods. The code is available at \url{https://github.com/JiahuaDong/FISS}.

\end{abstract}

\vspace{-5pt}
\section{Introduction}\label{sec:intro}
\vspace{-5pt}
Federated learning (FL) \cite{fed_average, model-agnostic, feserated_unsupervised, scaffold} is a remarkable decentralized training paradigm to learn a global model across distributed local clients without accessing their private data. Under privacy preservation, it has achieved rapid development in semantic segmentation \cite{deeplab, fullyconvolution, 9616392_Dong} by training on multiple decentralized local clients to alleviate the constraint of data island that requires enormous finely-labeled pixel annotations \cite{tumour_segmentation}. As a result, federated learning-based semantic segmentation (FSS) \cite{feddg, Federated_imitation} significantly economizes annotation costs in data-scarce scenarios via training a global segmentation model on private data of different clients \cite{feddg}.

\begin{figure}
    \centering
    \includegraphics[trim = 58mm 38mm 58mm 38mm, clip, width=240pt, height=190pt]{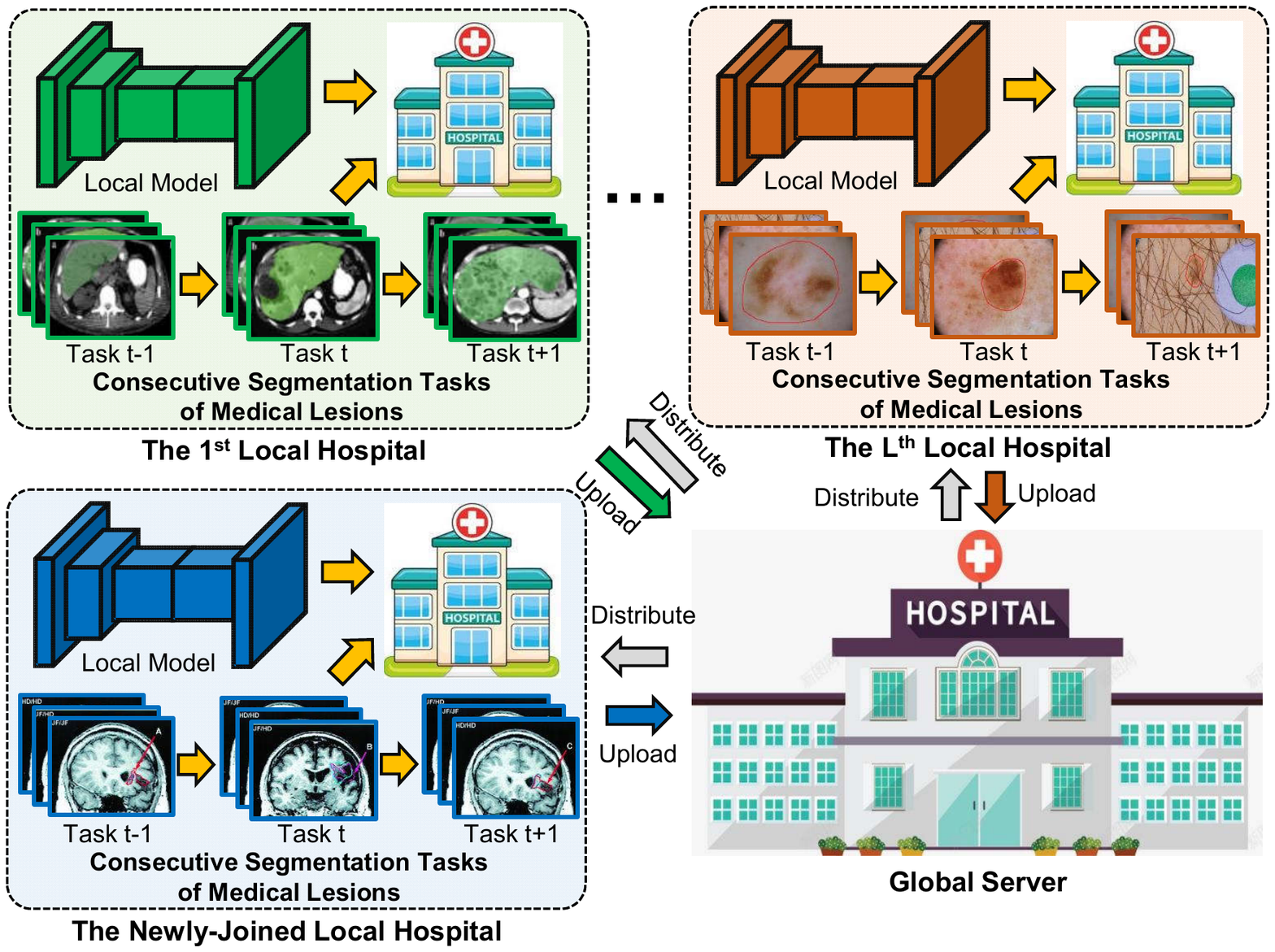}
    \vspace{-20pt}
    \caption{Exemplary FISS setting for medical diagnosis. Hundreds of hospitals including newly-joined ones receive new classes incrementally according to their own preference. FISS aims to segment new diseases consecutively via collaboratively learning a global segmentation model on private medical data of different hospitals.}
    \label{Fig:teaser}
    \vspace{-10pt}
\end{figure}

However, existing FSS methods \cite{feddg, Federated_imitation, feddrive2022, tumour_segmentation} unrealistically assume that the learned foreground classes are static and fixed over time, which is impractical in real-world dynamic applications where local clients receive streaming data of new categories consecutively. To tackle this issue, existing FSS methods \cite{Federated_imitation, MEHTA2022197, feddg} typically enforce local clients to store all samples of previously-learned old classes, and then learn a global model to segment new categories continually via FL. Nevertheless, it requires large computation and memory overhead as new classes arrive continuously, limiting the application ability of FSS methods \cite{Federated_imitation, feddrive2022}. If local clients have no memory to store old classes, existing FSS methods \cite{feddrive2022, feddg} significantly degrade segmentation behavior on old categories (\emph{i.e.}, catastrophic forgetting \cite{wei2022incremental, icarl, ijcai2020_77}) when learning new classes incrementally. In addition, the pixels labeled as background in the current learning task may belong to old classes from old tasks or new foreground classes from future tasks.
This phenomenon is also known as background shift \cite{PLOP, ILT} that heavily aggravates heterogeneous forgetting speeds on old categories. More importantly, in practical scenarios, new local clients receiving new categories incrementally may join in global FL training irregularly, thus further exacerbating catastrophic forgetting to some extent.

To surmount the above real-world scenarios, we propose a novel practical problem called {\textbf{F}}ederated {\textbf{I}}ncremental {\textbf{S}}emantic {\textbf{S}}egmentation (\textbf{FISS}), where local clients collect new categories consecutively according to their preferences, and new local clients collecting unseen novel classes participate in global FL training irregularly. In the FISS settings, the class distributions are non-independent and identically distributed (Non-IID) across different clients, and training data of old classes is unavailable for all local clients. FISS aims to train a global incremental segmentation model via collaborative FL training on local clients while addressing catastrophic forgetting. In this paper, we use medical lesions segmentation \cite{tumour_segmentation, feddg} as an example to better illustrate FISS, as shown in Figure~\ref{Fig:teaser}. Hundreds of hospitals, as well as newly joined ones, collect unseen/new medical lesions continuously in clinical diagnosis. Considering privacy preservation, it is desired for these hospitals to learn a global segmentation modal via FL without accessing each other's data~\cite{fed_average, zhang2022towards}.


A naive solution for FISS problem is to directly integrate incremental semantic segmentation \cite{MIB, PLOP, RCIL} and FL \cite{10.1145/3298981, 10.5555/2999611.2999748} together. Nevertheless, such a trivial solution requires global server to have strong human prior about which and when local clients can collect new categories, so that global model learned in the latest old task can be stored by local clients to address forgetting on old classes via knowledge distillation \cite{44873_Distilling, wang2022foster}. Considering privacy preservation in the FISS, this privacy-sensitive prior knowledge cannot be shared between local clients and global server.
As a result, this naive solution severely suffers from intra-client heterogeneous forgetting on different old classes caused by background shift \cite{MIB, PLOP, RCIL, ILT}, and inter-client heterogeneous forgetting across different clients brought by Non-IID class distributions.

To overcome the above-mentioned challenges, we develop a novel  \underline{\textbf{F}}orgetting-\underline{\textbf{B}}alanced \underline{\textbf{L}}earning (\textbf{FBL}) model, which alleviates heterogeneous forgetting on old classes from intra-client and inter-client perspectives. Specifically, to tackle intra-client heterogeneous forgetting caused by background shift, we propose an adaptive class-balanced pseudo labeling to adaptively generate confident pseudo labels for old classes. Under the guidance of pseudo labels, we propose a forgetting-balanced semantic compensation loss to rectify different forgetting of old classes with background shift via considering balanced gradient propagation of local clients. In addition, a forgetting-balanced relation consistency loss is designed to distill underlying category-relation consistency between old and new classes for intra-client heterogeneous forgetting compensation. Moreover, considering addressing heterogeneous forgetting from inter-client aspect, we develop a task transition monitor to automatically identify new classes without any human prior, and store the latest old model from global perspective for relation consistency distillation. Experiments on segmentation datasets reveal large improvement of our model over comparison methods. We summarize the main contributions of this work as follows:

\begin{myitemize}
	\item We propose a novel practical problem called Federated Incremental Semantic Segmentation (FISS), where the major challenges are intra-client and inter-client heterogeneous forgetting on old categories caused by intra-client background shift and inter-client Non-IID distributions.
	
	\item We propose a Forgetting-Balanced Learning (FBL) model to address the FISS problem via surmounting heterogeneous forgetting from both intra-client and inter-client aspects. As we all know, in the FL field, this is a pioneer attempt to explore a global continual segmentation model.
	
	\item We develop a forgetting-balanced semantic compensation loss and a forgetting-balanced relation consistency loss to tackle intra-client heterogeneous forgetting across old classes, under the guidance of confident pseudo labels generated via adaptive class-balanced pseudo labeling.
	
	\item We design a task transition monitor to surmount inter-client heterogeneous forgetting by accurately recognizing new classes under privacy protection and storing the latest old model from global aspect for relation distillation.
\end{myitemize}

\section{Related Work}\label{sec:related_works}
\vspace{-5pt}
\textbf{Federated Learning (FL)} \cite{fed_average, fed_speedup, fedbn, federated_aver} aggregates local-client model parameters to optimize a global model under privacy protection.  \cite{convergence} enforces local model to approximate the global ones via a proximal term. To minimize computation cost, \cite{layer-wise} employs a layer-wise parameter aggregation strategy. Inspired by above FL \cite{fed_average, fed_bayesian, model-agnostic} methods, \cite{tumour_segmentation, feddg, Federated_imitation} apply FL to semantic segmentation ~\cite{atrousconvolution, fullyconvolution}, which has achieved rapid developments in medical analysis \cite{MEHTA2022197, What_Transferred_Dong_CVPR2020} and autonomous driving \cite{feddrive2022}. \cite{DBLP:conf/iclr/PengHZS20} considers adversarial framework \cite{zhang2020principal, zhang2021causaladv} to tackle domain adaptation problem \cite{9186366, li2021DCC, wangnon, li2020content} in the FL field. 
\cite{dong2022federated} proposes a federated class-incremental learning model via considering global and local forgetting. However, the above-mentioned methods \cite{feddg, Daiqiong_qi, feddrive2022} cannot segment new foreground classes continuously under the FISS settings.





\textbf{Incremental Semantic Segmentation (ISS)} \cite{PLOP, RCIL, MIB, SDR} considers class-incremental learning \cite{EWC,DGR,GEM} in semantic segmentation. The key challenges of ISS are catastrophic forgetting \cite{catastrophic_forgetting, icarl} and background shift \cite{PLOP, SDR}, as claimed in \cite{MIB, PACKNET}. ILT~\cite{ILT} proposes to distill latent features and probabilities between old and new models. PLOP~\cite{PLOP} utilizes multi-scale pooling distillation to maintain past experience. SDR~\cite{SDR} considers feature consistency by prototype matching and contrastive learning \cite{DBLP:journals/ijautcomp/ChenZHCSXX23}. RCIL~\cite{RCIL} decouples the network into branches to overcome forgetting. Considering tackling background shift, \cite{PLOP, STISS, ssul} propose pseudo labeling to annotate old classes labeled as background pixels. 
Nevertheless, these ISS methods \cite{PLOP, STISS, RCIL} cannot be effectively applied to address the FISS problem, due to their strong  prior knowledge to access privately-sensitive information (\emph{i.e.}, when and which local clients receive new classes).



\section{Problem Definition}\label{sec: problem_definition}
\vspace{-5pt}
As claimed in incremental semantic segmentation (ISS) \cite{LWF, ILT, MIB, PLOP}, some consecutive segmentation tasks are defined as $\mathcal{T} = \{\mathcal{T}^t\}_{t=1}^T$, where the $t$-th ($t=1, \cdots, T$) task $\mathcal{T}^t = \{\mf{x}_i^t, \mf{y}_i^t\}_{i=1}^{N^t}$ is composed of $N^t$ pairs of RGB images $\mf{x}_i^t\in\mathbb{R}^{H\times W \times 3}$ and labels $\mf{y}_i^t\in\mathbb{R}^{H\times W}$. $H$ and $W$ denote height and width of given images. The label space $\mathcal{Y}^t$ of $t$-th incremental task consists of $K^t$ new categories and background. $K^t$ new classes have no overlap with $K^o=\sum_{i=1}^{t-1}K^i \subset \cup_{j=1}^{t-1}\mathcal{Y}^j$ old classes learned from $t-1$ old tasks. In the $t$-th task, we follow ISS methods \cite{ILT, PLOP} to annotate $K^o$ old classes and other foreground classes from future learning tasks as background (\emph{i.e.}, background shift \cite{MIB}), due to unavailable training data of $K^o$ old classes.

We then extend the settings of incremental semantic segmentation (ISS) \cite{ILT, MIB, RCIL} to Federated Incremental Semantic Segmentation (FISS). Denote global server as $\mathcal{S}_g$ and $L$ local clients as $\{\mathcal{S}_l\}_{l=1}^L$. In the FISS, at the $r$-th ($r=1, \cdots, R$) global round, we randomly select some local clients to aggregate gradients. When we choose the $l$-th local client to learn the $t$-th segmentation task, the latest global model $\Theta^{r, t}$ is distributed to $\mathcal{S}_l$, and trained on private training data $\mathcal{T}_l^t = \{\mf{x}_{li}^{t}, \mf{y}_{li}^t\}_{i=1}^{N_l^t}\sim\mathcal{P}_l$ of $\mathcal{S}_l$. $\mf{x}_{li}^{t}$ and $\mf{y}_{li}^t\in\mathcal{Y}_l^t$ denote the images and labels of the $l$-th client. $\{\mathcal{P}_l\}_{l=1}^L$ are non-independent and identically distributed (\emph{i.e.}, Non-IID) across local clients. The label space $\mathcal{Y}_l^t\subset \mathcal{Y}^t$ of $\mathcal{S}_l$ in the $t$-th task is composed of $K_l^t$ new classes ($K_l^t \leq K^t$) that belongs to a subset of $\mathcal{Y}^t=\cup_{l=1}^L\mathcal{Y}_l^t$. Following ISS methods \cite{ILT, PLOP, RCIL}, we consider background shift in the FISS and also annotate $K_l^o=\sum_{i=1}^{t-1}K_l^i\subset \cup_{j=1}^{t-1}\mathcal{Y}_l^j$ old categories from $t-1$ old tasks and other foreground categories from future learning tasks as background. After getting global model $\Theta^{r, t}$ and performing local training on $\mathcal{T}_l^t$, $\mathcal{S}_l$ obtains a updated local model $\Theta_{l}^{r, t}$. Then global server $\mathcal{S}_g$ aggregates local models of selected clients as the global model $\Theta^{r+1, t}$ for the training of next global round.

In the $t$-th task, motivated by \cite{dong2022federated}, all local clients $\{\mathcal{S}_l\}_{l=1}^L$ are divided into three categories: $\{\mathcal{S}_l\}_{l=1}^L = \mathbf{S}_o \cup \mathbf{S}_c \cup \mathbf{S}_n$. Specifically, $\mathbf{S}_o$ is composed of $L_o$ local clients that have accumulated past experience for previous tasks but cannot collect new training data of the $t$-th task; $\mathbf{S}_c$ consisting of $L_c$ local clients can receive new training data of current task and has learning experience for old classes; $\mathbf{S}_n$ includes $L_n$ new local clients with unseen novel classes but without past learning experience of old classes. These local clients are randomly determined in each incremental task. New clients $\mathbf{S}_n$ are added randomly at any global round in the FISS, increasing $L=L_o+L_c+L_n$ gradually as continuous tasks. More importantly, we don't have prior knowledge about the class distributions $\{\mathcal{P}_l\}_{l=1}^L$, quantity and order of segmentation tasks, when and which local clients receive new classes. In this paper, FISS aims to learn a global model $\Theta^{R, T}$ to segment new categories continuously while surmounting heterogeneous forgetting on old categories brought by background shift, under the privacy preservation of local clients.

\section{The Proposed Model}\label{sec: proposed_model}
\begin{figure*}[t]
	\centering
	\includegraphics[trim = 25mm 50mm 26mm 50mm, clip, width =495pt, height=215pt]
	{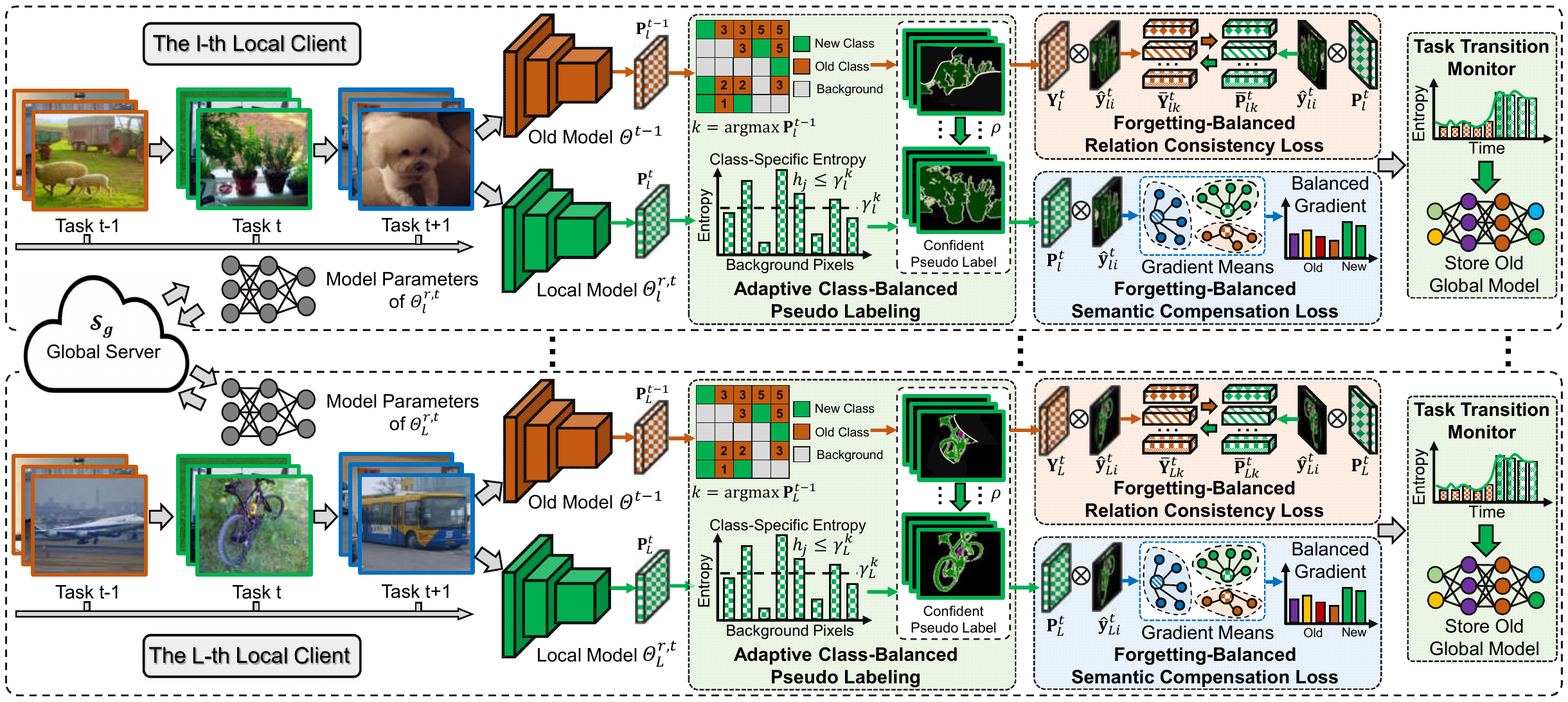}
	\vspace{-20pt}
	\caption{Overview of the proposed FBL model. It includes a \textit{forgetting-balanced semantic compensation loss} $\mathcal{L}_{\mr{FS}}$ and a \textit{forgetting-balanced relation consistency loss} $\mathcal{L}_{\mr{FR}}$ to tackle intra-client heterogeneous forgetting brought by background shift, under the guidance of \textit{adaptive class-balanced pseudo labeling}. Meanwhile, it utilizes a \textit{task transition monitor} to overcome inter-client heterogeneous forgetting brought by Non-IID distributions with background shift. }
	\label{fig: overview_of_our_model}
	\vspace{-10pt}
\end{figure*}

\vspace{-5pt}
Figure~\ref{fig: overview_of_our_model} presents the overview of our model to address the FISS problem. Our FBL model overcomes intra-client heterogeneous forgetting via a forgetting-balanced semantic compensation loss (Section \ref{sec: semantic_compensation}) and a forgetting-balanced relation consistency loss (Section \ref{sec: relation_consistency}), under the guidance of adaptive class-balanced pseudo labeling (Section \ref{sec: pseudo_labeling}) to mine pseudo labels for old classes with background shift. Meanwhile, it addresses inter-client heterogeneous forgetting via a task transition monitor (Section \ref{sec: task_transition}) to recognize new classes and store old model for relation distillation.

\subsection{Adaptive Class-Balanced Pseudo Labeling}\label{sec: pseudo_labeling}

For the $l$-th local client $\mathcal{S}_l\in\mathbf{S}_c\cup\mathbf{S}_n$, the semantic segmentation loss $\mathcal{L}_{\mr{SE}}$ for a mini-batch $\{\mf{x}_{li}^t, \mf{y}_{li}^t\}_{i=1}^B\subset \mathcal{T}_l^t$ sampled from the $t$-th incremental task is formulated as:
\begin{align}\small
		\mathcal{L}_{\mr{SE}} =\frac{1}{B}\sum_{i=1}^B\sum_{j=1}^{HW} \mathcal{D}_{\mr{CE}}\big(\mathbf{P}_l^t(\mathbf{x}_{li}^t, \Theta^{r, t})_j, (\mathbf{y}_{li}^t)_j\big), 
		\label{eq: classification_loss}
\end{align}
where $\mathcal{D}_{\mr{CE}}(\cdot, \cdot)$ denotes the cross-entropy loss. At the $r$-th global round, global model $\Theta^{r, t}$ is transmitted from global server $\mathcal{S}_g$ to $\mathcal{S}_l$. $\mathbf{P}_l^t(\mathbf{x}_{li}^t, \Theta^{r, t})_j\in\mathbb{R}^{1+K^o+K^t}$ is the probability at the $j$-th ($j=1, \cdots, HW$) pixel predicted by $\Theta^{r, t}$, and it predicts background, $K^o$ old classes, and $K^t$ new classes for the $j$-th pixel. $(\mathbf{y}_{li}^t)_j\in\mathcal{Y}_l^t$ is corresponding label at the $j$-th pixel. $B$ represents the batch size.

As aforementioned, in the FISS settings, local client $\mathcal{S}_l$ has no memory to store $K^o$ old classes, while background pixels may belong to $K^o$ old classes, other foreground classes from future tasks or real background (\emph{i.e.}, background shift \cite{MIB, PLOP}). As a result, it enforces the updating of local model $\Theta_{l}^{r, t}$ (\emph{i.e.}, Eq.~\eqref{eq: classification_loss}) to suffer from intra-client heterogeneous forgetting among different old classes brought by background shift, after $\mathcal{S}_l$ receives the global model $\Theta^{r, t}$ from $\mathcal{S}_g$ for local training. 
To this end, as shown in Figure~\ref{fig: overview_of_our_model}, we develop an adaptive class-balanced pseudo labeling to adaptively mine confident pseudo labels for old classes labeled as background pixels in the $t$-th segmentation task. Different from existing ISS methods \cite{PLOP, STISS, ssul} that only utilize a constant probability threshold to select pseudo labels for all classes, our FBL model considers class balance to mine pseudo labels for old classes via introducing class-specific entropy threshold for each old class, which are determined as continual learning process. These class-balanced pseudo labels of $K^o$ old classes are essential to alleviate heterogeneous forgetting of old classes.


In the $t$-th task, as shown in Figure~\ref{fig: overview_of_our_model}, given a sample $\{\mf{x}_{li}^t, \mf{y}_{li}^t\}\subset \mathcal{T}_l^t$, we feed it into old global model $\Theta^{t-1}$ of the last task and current local model $\Theta_{l}^{r, t}$ to obtain the probabilities $\mathbf{P}_l^{t-1}(\mf{x}_{li}^t, \Theta^{t-1})\in\mathbb{R}^{H\times W\times(1+K^o)}$ and $\mathbf{P}_l^{t}(\mf{x}_{li}^t, \Theta_l^{r, t})\in\mathbb{R}^{H\times W\times(1+K^o+K^t)}$ respectively. Then pseudo label $\hat{\mathbf{y}}_{li}^t\in\mathbb{R}^{H\times W}$ of given image $\mf{x}_{li}^t$ is defined as:
\begin{equation}\small
	\begin{split}
		(\hat{\mathbf{y}}_{li}^t)_j =
		\left\{
		\begin{aligned}		
			&k, ~\mr{if}~(\mathbf{y}_{li}^t)_j\notin \mathcal{Y}_l^b ~\mr{and}~ k=(\mathbf{y}_{li}^t)_j; \\
			&k, ~\mr{if}~(\mathbf{y}_{li}^t)_j\in \mathcal{Y}_l^b ~\mr{and}~ h_j\leq\gamma_l^k  \\
			& ~\quad~\mr{and}~ k=\arg\max \mathbf{P}_l^{t-1}(\mathbf{x}_{li}^t, \Theta^{t-1})_j; \\
			&0, ~\mr{otherwise}, \\
		\end{aligned} 					
		\right.											
	\end{split}							
	\label{eq: pseudo_label}	
\end{equation}
where $(\hat{\mathbf{y}}_{li}^t)_j$ is pseudo label of the $j$-th pixel from $\hat{\mathbf{y}}_{li}^t$. $\mathbf{P}_l^{t-1}(\mf{x}_{li}^t, \Theta^{t-1})_j$ is softmax probability of the $j$-th pixel from $\mathbf{P}_l^{t-1}(\mf{x}_{li}^t, \Theta^{t-1})$. $h_j=\mathcal{H}(\mathbf{P}_l^{t}(\mf{x}_{li}^t, \Theta_l^{r, t})_j)$ represents entropy of the $j$-th pixel, and $\mathcal{H}(\mathbf{p}) = \sum_i\mathbf{p}_i\log\mathbf{p}_i$ is entropy measure function. $\{\gamma_l^k\}_{k=1}^{K^o}$ denote class-specific entropy threshold to adaptively select class-balanced pseudo labels with high confidence. As shown in Eq.~\eqref{eq: pseudo_label}, in the $t$-th task $\mathcal{T}_l^t$, when the $j$-th pixel belongs to background label space $\mathcal{Y}_l^b$ (\emph{i.e.}, $(\mathbf{y}_{li}^t)_j\in \mathcal{Y}_l^b$) and the entropy $h_j$ is less than $\gamma_l^k$, pseudo label is determined by $(\hat{\mathbf{y}}_{li}^t)_j = \arg\max \mathbf{P}_l^{t-1}(\mf{x}_{li}^t, \Theta^{t-1})_j$. If the $j$-th pixel is not labeled as background (\emph{i.e.}, $(\mathbf{y}_{li}^t)_j\notin \mathcal{Y}_l^b$), we consider pseudo label as new foreground classes: $(\hat{\mathbf{y}}_{li}^t)_j=(\mathbf{y}_{li}^t)_j$. Otherwise, $(\hat{\mathbf{y}}_{li}^t)_j=0$ denotes real background for the $j$-th pixel of $\hat{\mathbf{y}}_{li}^t$.

\setlength{\textfloatsep}{5pt}
\begin{algorithm}[!t]
	\footnotesize
	\caption{Determination of $\{\gamma_l^k\}_{k=1}^{K^o}$ in Eq.~\eqref{eq: pseudo_label}. }
	\label{alg: determinater_lambda}
	\LinesNumbered 
	\textbf{Input:} $\mathcal{T}_l^t = \{\mf{x}_{li}^{t}, \mf{y}_{li}^t\}_{i=1}^{N_l^t}$, and the selection proportion $\rho$; 
	
	\For{$i=1, \cdots, N_l^t$}{
		$\mathbf{H}_{li}^t= \mathcal{H}(\mathbf{P}_l^{t}(\mf{x}_{li}^t, \Theta_l^{r, t}))\in\mathbb{R}^{H\times W}$;\\
		$\mathbf{L}_{li}^t=\arg\max \mathbf{P}_l^{t-1}(\mf{x}_{li}^t, \Theta^{t-1})\in\mathbb{R}^{H\times W}$;\\
		
		\For{$k=1, \cdots, K^o$}{
			$\mathbf{H}_l^k = \mathbf{H}_{li}^t[\mathbf{L}_{li}^t==k]$; \\
			$\mathbf{M}_l^k = [\mathbf{M}_l^k; \mr{matrix\_to\_vector}(\mathbf{H}_l^k)]$;  \\
		}
	}
	\For{$k=1, \cdots, K^o$}{
		$\mathbf{E}_l^k = \mr{sort}(\mathbf{M}_l^k, \mr{order=ascending})$; \\
		$\gamma_l^k = \mathbf{E}_l^k[\mr{length}(\mathbf{E}_l^k) \cdot \rho]$. \\
	}
\end{algorithm}

The determination of $\{\gamma_l^k\}_{k=1}^{K^o}$ is summarized in \textbf{Algorithm} \ref{alg: determinater_lambda}. After computing entropy $\{\mathbf{H}_{li}^t\}_{i=1}^{N_l^t}$ for all samples in the $t$-th task $\mathcal{T}_l^t$, we sort the entropy of all pixels predicted as the $k$-th class. $\gamma_l^k$ is determined via the entropy ranked at $[\mr{length}(\mathbf{E}_l^k)\cdot\rho]$ of $\mathbf{E}_l^k$, where $\rho$ is selection proportion for all old classes. The value of $\rho$ is initialized as $20\%$, and adds $10\%$ for each epoch empirically as training process. We set the maximum selection proportion $\rho$ as $80\%$. Given a mini-batch $\{\mf{x}_{li}^t, \mf{y}_{li}^t\}_{i=1}^B\subset \mathcal{T}_l^t$, we generate class-balanced pseudo labels $\{\mf{x}_{li}^t, \hat{\mf{y}}_{li}^t\}_{i=1}^B\subset \mathcal{T}_l^t$ adaptively via considering class-balanced selection proportion $\rho$ in Eq.~\eqref{eq: pseudo_label} for all old classes. These confident pseudo labels provide strong guidance for the forgetting-balanced semantic compensation loss (Section \ref{sec: semantic_compensation}) and forgetting-balanced relation consistency loss (Section \ref{sec: relation_consistency}) to surmount intra-client heterogeneous forgetting among different old classes.

\subsection{Forgetting-Balanced Semantic Compensation}\label{sec: semantic_compensation}
To address heterogeneous forgetting speeds of different old classes within local client $\mathcal{S}_l\in\mathbf{S}_c\cup\mathbf{S}_n$, we propose a forgetting-balanced semantic compensation loss $\mathcal{L}_{\mr{FS}}$, as shown in Figure~\ref{fig: overview_of_our_model}. It considers balanced gradient propagation between different old tasks for intra-client heterogeneous forgetting compensation. Specifically, the loss $\mathcal{L}_{\mr{FS}}$ employs gradient propagation means of different old tasks to measure the forgetting heterogeneity of old classes, and then reweights segmentation loss $\mathcal{L}_{\mr{SE}}$ in Eq.~\eqref{eq: classification_loss} to normalize heterogeneous forgetting speeds brought by background shift. For a given sample $\{\mf{x}_{li}^t, \hat{\mf{y}}_{li}^t\}\subset \mathcal{T}_l^t$ with generated pseudo label, we first obtain its probability $\mathbf{P}_l^t(\mathbf{x}_{li}^t, \Theta_l^{r, t})$ predicted via local model $\Theta_l^{r, t}$. Motivated by \cite{wang2021addressing}, we then formulate gradient scalar $\Gamma_{ij}^t$ of the $j$-th pixel with respect to the $k$-th output neuron $\mathcal{N}_{k}^t$ of pixel classifier in $\Theta_l^{r, t}$ as follows:
\begin{align}
	\!\!\Gamma_{ij}^t\!\! =\!\! \frac{\partial\mathcal{D}_{\mr{CE}}(\mathbf{P}_l^t(\mathbf{x}_{li}^t, \Theta_l^{r, t})_j, (\hat{\mathbf{y}}_{li}^t)_j)}{\partial\mathcal{N}_k^t} \!\! =\!\! \mathbf{P}_l^t(\mf{x}_{li}^t, \Theta_l^{r, t})_j^k \!-\! 1,\!\!
	\label{eq: gradient_value}
\end{align}
where $\mathbf{P}_l^t(\mf{x}_{li}^t, \Theta_l^{r, t})_j^k$ is probability of the $k$-th class at the $j$-th pixel of $\mathbf{x}_{li}^t$, and $k=(\hat{\mathbf{y}}_{li}^t)_j$ denotes pseudo label of the $j$-th pixel in $\mathbf{x}_{li}^t$. Considering that intra-client heterogeneous forgetting of old classes changes dynamically as continual learning tasks, we expect gradient scalar $\Gamma_{ij}^t$ of old classes to be adaptive in the FISS, and reformulate Eq.~\eqref{eq: gradient_value} as:
\begin{align}
	\!\bar{\Gamma}_{ij}^t \!\!=\!\! |\Gamma_{ij}^t|^{\frac{K_l^o}{K_l^o+K_l^t}}\!\cdot\! \mathbb{I}_{(\hat{\mf{y}}_{li}^t)_j\in\cup_{\eta=1}^{t-1}\mathcal{Y}_l^\eta} \!+\!  |\Gamma_{ij}^t|\!\cdot\!\mathbb{I}_{(\hat{\mf{y}}_{li}^t)_j\in\mathcal{Y}_l^t\cup\mathcal{Y}_l^b}. \!\!
	\label{eq: gradient_value_adaptive}
\end{align}
where $\mathcal{Y}_l^b$ is background label space of the $l$-th local client $\mathcal{S}_l$. When pseudo label $(\hat{\mathbf{y}}_{li}^t)_j$ of the $j$-th pixel in $\mathbf{x}_{li}^t$ belongs to old classes from previous $t\!-\!1$ tasks, $\bar{\Gamma}_{ij}^t = |\Gamma_{ij}^t|^{{K_l^o}/{(K_l^o+K_l^t)}}$; otherwise, $\bar{\Gamma}_{ij}^t=|\Gamma_{ij}^t|$ for new classes and background.

As a result, given mini-batch samples $\{\mathbf{x}_{li}^t, \hat{\mathbf{y}}_{li}^t\}_{i=1}^B\in\mathcal{T}_l^t$ in the $t$-th segmentation task, we denote gradient propagation means $\Gamma_{b}$ and $\Gamma_\eta$ for the background and foreground classes learned from the $\eta$-th ($1\leq\eta\leq t$) task as follows: 
\begin{align}
\Gamma_{b} = \frac{1}{Z_b} \sum_{i=1}^B\sum_{j=1}^{HW} \bar{\Gamma}_{ij}^t,~~
	\Gamma_\eta = \frac{1}{Z_\eta}\sum_{i=1}^B\sum_{j=1}^{HW} \bar{\Gamma}_{ij}^t,
	\label{eq: task_specific_gradient}
\end{align}
where the quantity of pixels belonging to background and the $\eta$-th task are denoted as $Z_b=\sum\nolimits_{i=1}^B \sum_{j=1}^{HW} \mathbb{I}_{(\hat{\mf{y}}_{li}^t)_j\in\mathcal{Y}_l^b}$ and $Z_\eta\!=\!\sum\nolimits_{i=1}^B\sum_{j=1}^{HW} \mathbb{I}_{(\hat{\mf{y}}_{li}^t)_j \in\mathcal{Y}_l^\eta}$. The gradient propagation means $\Gamma_b$ and $\{\Gamma_\eta\}_{\eta=1}^{t}$ in Eq.~\eqref{eq: task_specific_gradient} reflect gradient-imbalanced propagation between old and new classes. Thus, these gradient means can effectively measure intra-client forgetting heterogeneity of different old classes, and evaluate updating speeds of new classes and background to some extent. Under the guidance of pseudo labels $\hat{\mathbf{y}}_{li}^t$, we employ $\{\Gamma_{\eta}\}_{\eta=1}^{t}$ and $\Gamma_{b}$ to reweight semantic segmentation loss $\mathcal{L}_{\mr{SE}}$, and formulate the proposed forgetting-balanced semantic compensation loss $\mathcal{L}_{\mr{FS}}$ as follows:
\begin{align}\setlength{\jot}{0pt}
	\mathcal{L}_{\mr{FS}} \!=\!\frac{1}{B}\sum_{i=1}^B\sum_{j=1}^{HW}\frac{\bar{\Gamma}_{ij}^t}{\bar{\Gamma}} \cdot \mathcal{D}_{\mr{CE}}\big(\mathbf{P}_l^t(\mathbf{x}_{li}^t, \Theta_l^{r, t})_j, (\hat{\mathbf{y}}_{li}^t)_j\big), 
	\label{eq: forgetting_balanced_loss}
\end{align}
where $\bar{\Gamma} = \sum_{\eta=1}^{t}	\Gamma_\eta \cdot \mathbb{I}_{(\hat{\mf{y}}_{li}^t)_j \in\mathcal{Y}_l^\eta} + \Gamma_{b} \cdot \mathbb{I}_{(\hat{\mf{y}}_{li}^t)_j \in\mathcal{Y}_l^b}$ denotes different normalization weights for background, old and new classes. $\mathcal{L}_{\mr{FS}}$ can address intra-client heterogeneous forgetting of different old classes via reweighting segmentation loss $\mathcal{L}_{\mr{SE}}$ to achieve class-balanced gradient propagation.

\subsection{Forgetting-Balanced Relation Consistency}\label{sec: relation_consistency}
The intrinsic relations between old and new classes are immutable in purely semantic space, independent of background shift \cite{MIB, PLOP} and availability of training data of old classes. In light of this, consistent semantic relations between old model $\Theta^{t-1}$ and current local model $\Theta_{l}^{r, t}$ plays an important role in tackling intra-client heterogeneous forgetting on old classes. However, most existing ISS methods \cite{LWF, MIB, ILT} only consider underlying relationships among old classes via performing knowledge distillation \cite{44873_Distilling} on an individual sample, which can be severely affected by noisy predictions on old classes brought by background shift. In addition, forgetting heterogeneity of old classes within local clients enforces most ISS methods \cite{PLOP, MIB} to suffer from heterogeneous inter-class relations distillation, thus aggravating imbalanced gradient propagation across incremental tasks.

\begin{table*}[t]
\centering
\setlength{\tabcolsep}{1.0mm}
\caption{Comparisons of mIoU (\%) on Pascal-VOC 2012 dataset \cite{10.1007/s11263-009-0275-4} under the setting of 15-1 with overlapped foregrounds. }
\vspace{-10pt}
\resizebox{\linewidth}{!}{
	\begin{tabular}{c|ccccccccccccccccccccc|>{\columncolor{lightgray}}c|>{\columncolor{lightgray}}c}
		\toprule
		Class ID &0& 1 & 2 & 3 & 4 & 5 &6 & 7 & 8 & 9 & 10 & 11 & 12 & 13 & 14 & 15 & 16 & 17 & 18 & 19 & 20 & mIoU & Imp.  \\
		\midrule
  
		Finetuning + FL & 70.7 & 6.5 & 0.0 & 0.0 & 11.2 & 0.1 & 0.9 & 0.1 & 0.0 & 0.0 & 0.0 & 0.0 & 0.0 & 0.0 & 0.0 & 0.0 & 0.0& 0.0 & 0.0 & 0.0 & 9.7 & 4.7  & $\Uparrow$ 51.9 \\

		LWF \cite{LWF} + FL & 81.6 & 0.2 & 0.0 & 0.0 & 8.9 & 0.0 & 0.0 & 0.0 & 0.0 & 0.0 & 0.0 & 0.0 & 0.0 & 0.0 & 0.0 & 2.9 & 0.0& 0.0 & 0.0 & 7.2 & 8.6 & 5.2  & $\Uparrow$ 51.4 \\
  
		ILT \cite{ILT}  + FL & 82.3 & 13.1 & 0.0 & 0.0 & 8.2 & 0.0 & 5.5 & 0.0 & 0.0 & 3.2 & 0.3 & 11.3 & 0.0 & 17.6 & 0.1 & 1.8 & 0.0& 0.0 & 5.7 & 7.8 & 13.0 & 8.1  & $\Uparrow$ 48.5 \\

		MiB \cite{MIB} + FL  & \textcolor{blue}{\textbf{84.9}} & 15.9 & 31.7 & 35.8 & 17.9 & 37.8 & 9.1 & 47.2 & 62.9 & 10.6 & 42.2 & 25.5 & 54.7 & 48.3 & 50.8 & \textcolor{blue}{\textbf{77.7}} & 0.0& 6.2 & 8.1 & 15.8 & 13.2 & 33.1  & $\Uparrow$ 23.5 \\

		PLOP \cite{PLOP} + FL & 62.7 & 55.1 & 20.0 & \textcolor{blue}{\textbf{49.6}} & 44.3 & \textcolor{blue}{\textbf{60.1}} & \textcolor{blue}{\textbf{82.4}} & 61.4 & 74.5 & 24.2 & \textcolor{blue}{\textbf{43.7}} & 43.9 & 57.6 & 48.3 & 61.2 & 67.3 & \textcolor{deepred}{\textbf{14.6}} & \textcolor{deepred}{\textbf{44.4}} & \textcolor{blue}{\textbf{10.4}} & 22.9 & 8.0 & 45.5  & $\Uparrow$ 11.1 \\

		RCIL \cite{RCIL} + FL & 0.0 & \textcolor{blue}{\textbf{76.0}} & \textcolor{deepred}{\textbf{41.9}} & 49.2 & \textcolor{deepred}{\textbf{63.4}} & 56.9 & \textcolor{deepred}{\textbf{84.2}} & \textcolor{deepred}{\textbf{82.5}} & \textcolor{deepred}{\textbf{85.3}} & \textcolor{deepred}{\textbf{36.5}} & 17.0 & \textcolor{blue}{\textbf{55.7}} & \textcolor{deepred}{\textbf{74.6}} & \textcolor{blue}{\textbf{64.2}} & \textcolor{blue}{\textbf{78.8}} & 68.2 & 0.9& \textcolor{blue}{\textbf{29.0}} & \textcolor{deepred}{\textbf{15.3}} & \textcolor{deepred}{\textbf{43.0}} & \textcolor{deepred}{\textbf{28.3}} & \textcolor{blue}{\textbf{50.0}}  & $\Uparrow$ 6.6 \\

		\midrule
		\textbf{FBL} (Ours) & \textcolor{deepred}{\textbf{88.7}} & \textcolor{deepred}{\textbf{81.9}} & \textcolor{blue}{\textbf{37.3}} & \textcolor{deepred}{\textbf{79.1}} & \textcolor{blue}{\textbf{60.5}} & \textcolor{deepred}{\textbf{71.3}} & 81.9 & \textcolor{blue}{\textbf{79.7}} & \textcolor{blue}{\textbf{81.9}} & \textcolor{blue}{\textbf{34.6}} & \textcolor{deepred}{\textbf{58.3}} & \textcolor{deepred}{\textbf{57.0}} & \textcolor{blue}{\textbf{70.3}} & \textcolor{deepred}{\textbf{70.4}} & \textcolor{deepred}{\textbf{79.4}} & \textcolor{deepred}{\textbf{80.5}} & \textcolor{blue}{\textbf{1.8}} & 9.0 & 1.5 & \textcolor{blue}{\textbf{40.5}} & \textcolor{blue}{\textbf{23.6}} & \textcolor{deepred}{\textbf{56.6}}  &  -- \\
		\bottomrule
\end{tabular}}
\label{tab: comparison_voc_15_1}
\vspace{-7pt}
\end{table*}

\begin{table*}[t]
\centering
\setlength{\tabcolsep}{1.05mm}
\caption{Comparisons of mIoU (\%) on Pascal-VOC 2012 dataset \cite{10.1007/s11263-009-0275-4} under the setting of 4-4 with overlapped foregrounds. }
\vspace{-10pt}
	\resizebox{\linewidth}{!}{
	\begin{tabular}{c|ccccccccccccccccccccc|>{\columncolor{lightgray}}c|>{\columncolor{lightgray}}c}
		\toprule
		Class ID & 0&1 & 2 & 3 & 4 & 5 &6 & 7 & 8 & 9 & 10 & 11 & 12 & 13 & 14 & 15 & 16 & 17 & 18 & 19 & 20 & mIoU& Imp.  \\
		\midrule

		Finetuning + FL & 73.1 & 0.0 & 0.0 & 0.0 & 0.0 & 0.0 & 0.0 & 0.0 & 0.0 & 0.0 & 0.0 &0.0& 0.0 &  0.0 & 0.5 & 0.0 & 0.0 & 20.1 & \textcolor{deepred}{\textbf{34.4}} & \textcolor{blue}{\textbf{32.3}} & 30.4  & 9.1&$\Uparrow$ 34.8 \\

		LWF \cite{LWF} + FL & \textcolor{deepred}{\textbf{88.4}} & 0.0 & 0.0 & 0.0 & 0.9 & 0.0 & 0.0 & 8.7 & 16.3 & \textcolor{deepred}{\textbf{8.5}} & 0.0 &\textcolor{deepred}{\textbf{39.2}}& \textcolor{blue}{\textbf{39.6}} &  \textcolor{blue}{\textbf{38.6}} & 63.2 & \textcolor{deepred}{\textbf{77.7}} & \textcolor{deepred}{\textbf{24.9}} & 15.1 & 24.9 & 25.1 & 29.6  & 23.8&$\Uparrow$ 20.1 \\

		ILT \cite{ILT}  + FL & \textcolor{blue}{\textbf{87.8}} & 0.0 & 0.0 & 0.0 & 8.5 & 0.1 & 4.1 & 22.7 & 14.5 & 2.4 & 0.0 &\textcolor{blue}{\textbf{34.5}}& 25.2 &  36.0 & 63.5 & \textcolor{blue}{\textbf{74.4}} & \textcolor{blue}{\textbf{15.2}} & 13.5 & 23.4 & 24.7 & 26.0  & 22.7&$\Uparrow$ 21.2 \\

		MiB \cite{MIB} + FL  & 86.7 & 50.9 & 23.0 & 17.7 & 25.0 & \textcolor{deepred}{\textbf{8.9}} & 41.3 & \textcolor{blue}{\textbf{67.1}} & \textcolor{blue}{\textbf{47.9}} & \textcolor{blue}{\textbf{4.5}} & 0.1 &29.1& 26.9 &  21.7 & \textcolor{deepred}{\textbf{69.8}} & 73.2 & 3.1 & 17.9 & \textcolor{blue}{\textbf{30.3}} & 29.2 & 19.8  & \textcolor{blue}{\textbf{33.0}}&$\Uparrow$ 10.9 \\

		PLOP \cite{PLOP} + FL & 85.5 & 1.7 & 0.3 & 0.0 &\textcolor{deepred}{\textbf{44.3}} & 0.2 & \textcolor{deepred}{\textbf{66.1}} & 58.1 & 0.6 & 0.0 & 1.9 &25.1& 33.4 &  31.0 & 46.1 & 70.3 & 0.0 & \textcolor{blue}{\textbf{27.5}} & 25.0 & \textcolor{deepred}{\textbf{36.1}} & 36.5  & 28.1&$\Uparrow$ 15.8 \\

		RCIL \cite{RCIL} + FL & 85.6 & \textcolor{blue}{\textbf{62.8}} & \textcolor{blue}{\textbf{29.6}} & \textcolor{blue}{\textbf{38.9}} &\textcolor{blue}{\textbf{39.3}} & 0.9 & 62.3 & 51.2 & 32.6 & 0.3 & \textcolor{deepred}{\textbf{34.1}} &21.1& 3.9 &  18.1 & 40.8 & 68.6 & 1.2 & 6.5 & 27.7 & 15.0 & \textcolor{blue}{\textbf{39.1}}  & 32.4&$\Uparrow$ 11.5 \\

		\midrule
		\textbf{FBL} (Ours)  & 86.3 & \textcolor{deepred}{\textbf{66.2}} & \textcolor{deepred}{\textbf{34.0}} & \textcolor{deepred}{\textbf{48.3}} &28.0 & \textcolor{deepred}{\textbf{6.9}} & \textcolor{blue}{\textbf{64.7}} & \textcolor{deepred}{\textbf{75.6}} & \textcolor{deepred}{\textbf{74.1}} & 0.0 & \textcolor{blue}{\textbf{26.0}} &29.9& \textcolor{deepred}{\textbf{61.7}} &  \textcolor{deepred}{\textbf{40.1}} & \textcolor{blue}{\textbf{66.0}} & 70.4 & 0.0 & \textcolor{deepred}{\textbf{40.4}} & 27.4 & 26.8 & \textcolor{deepred}{\textbf{48.5}}  & \textcolor{deepred}{\textbf{43.9}}&-- \\
  
		\bottomrule
\end{tabular}}
\label{tab: comparison_voc_4_4}
\vspace{-10pt}
\end{table*}

To this end, we propose a forgetting-balanced relation consistency loss $\mathcal{L}_{\mr{FR}}$ to tackle intra-client heterogeneous forgetting via compensating heterogeneous relation distillation. Specifically, we propose relationship prototype of each class instead of an individual sample to better characterize underlying relations between old and new classes, and consider gradient means in Eq.~\eqref{eq: task_specific_gradient} to balance heterogeneous relation distillation. As shown in Figure~\ref{fig: overview_of_our_model}, given $\{\mf{x}_{li}^t, \hat{\mf{y}}_{li}^t\}_{i=1}^B\subset \mathcal{T}_l^t$, we feed it into old model $\Theta^{t-1}$ and local model $\Theta_l^{r, t}$ of $\mathcal{S}_l$ to obtain probabilities $\mathbf{P}_l^{t-1}(\mathbf{x}_{li}^t, \Theta^{t-1})\in\mathbb{R}^{H\times W\times (1+K^o)}$ and $\mathbf{P}_l^t(\mathbf{x}_{li}^t, \Theta_l^{r, t})\in\mathbb{R}^{H\times W\times (1+K^o+K^t)}$. Then we substitute the first $1+K^o$ channel dimensions of one-hot pseudo label $\hat{\mf{Y}}_{li}^t\in\mathbb{R}^{H\times W\times (1+K^o+K^t)}$ ($\hat{\mf{Y}}_{li}^t$ is one-hot encoding of $\hat{\mf{y}}_{li}^t$) with $\mathbf{P}_l^{t-1}(\mathbf{x}_{li}^t, \Theta^{t-1})$, and abbreviate this variant as relationship label $\mathbf{Y}_{l}^t(\mathbf{x}_{li}^t, \Theta^{t-1})\in\mathbb{R}^{H\times W\times (1+K^o+K^t)}$ indicating underlying relations among old and new categories. For the $k$-th class, the relationship prototype $\bar{\mathbf{P}}_{lk}^t$ and its label $\bar{\mathbf{Y}}_{lk}^t$ are written as follows:
\begin{align}
	\bar{\mathbf{P}}_{lk}^t &= \frac{1}{Z_k}\sum_{i=1}^B\sum_{j=1}^{HW} \mathbf{P}_l^t(\mathbf{x}_{li}^t, \Theta_l^{r, t})_j\cdot \mathbb{I}_{(\hat{\mf{y}}_{li}^t)_j=k}, \\
	\bar{\mathbf{Y}}_{lk}^t &= \frac{1}{Z_k}\sum_{i=1}^B\sum_{j=1}^{HW} \mathbf{Y}_l^t(\mathbf{x}_{li}^t, \Theta^{t-1})\cdot \mathbb{I}_{(\hat{\mathbf{y}}_{li}^t)_j=k},
	\label{eq: relation_prototype}
\end{align}
where $Z_k = \sum_{i=1}^B\sum_{j=1}^{HW} \mathbb{I}_{(\hat{\mf{y}}_{li}^t)_j=k}$ is pixel number of the $k$-th class. The class-wise gradient mean $\Gamma_k$ for the $k$-th class is formulated as $\Gamma_k = \frac{1}{Z_k} \sum_{i=1}^B\sum_{j=1}^{HW} \bar{\Gamma}_{ij}^t\cdot \mathbb{I}_{(\hat{\mf{y}}_{li}^t)_j=k}$, which is then used to reweight heterogeneous distillation gains. As a result, the forgetting-balanced relation consistency loss $\mathcal{L}_{\mr{FR}}$ is concretely written as follows:
\begin{align}\small	
	\mathcal{L}_{\mr{FR}} = \frac{1}{K^o+K^t}\sum_{k=1}^{K^o+K^t} \frac{\Gamma_k}{\bar{\Gamma}_{\mr{cls}}} \cdot \mathcal{D}_{\mr{KL}}(\bar{\mathbf{P}}_{lk}^t, \bar{\mathbf{Y}}_{lk}^t),
	\label{eq: class_relation_distillation}
\end{align}
where $\mathcal{D}_{\mr{KL}}(\cdot||\cdot)$ is Kullback-Leibler divergence. $\bar{\Gamma}_{\mr{cls}} = \sum_{\eta=1}^{t}	\Gamma_\eta \cdot \mathbb{I}_{k\in\mathcal{Y}_l^\eta}$ denotes gradient normalization mean.

\begin{table*}[htb]
\centering
\setlength{\tabcolsep}{1.05mm}
\caption{Comparisons of mIoU (\%) on Pascal-VOC 2012 dataset \cite{10.1007/s11263-009-0275-4} under the setting of 8-2 with overlapped foregrounds. }
\vspace{-10pt}
\resizebox{\linewidth}{!}{
	\begin{tabular}{c|ccccccccccccccccccccc|>{\columncolor{lightgray}}c|>{\columncolor{lightgray}}c}
		\toprule
		Class ID &0& 1 & 2 & 3 & 4 & 5 &6 & 7 & 8 & 9 & 10 & 11 & 12 & 13 & 14 & 15 & 16 & 17 & 18 & 19 & 20 & mIoU& Imp.  \\
		\midrule

		Finetuning + FL & 70.1 & 0.0 & 0.0 & 0.0 & 0.0 & 0.0 & 0.0 & 0.0 & 0.0 & 0.0 & 0.0 & 0.0 & 0.0 & 0.0 & 0.0 & 0.0 & 0.0 & 0.0 & 0.0 & 0.0 & 13.6 & 4.0 & $\Uparrow$  31.7\\

		LWF \cite{LWF} + FL & \textcolor{blue}{\textbf{83.1}} & 0.2 & 0.0 & 0.0 & 5.0 & 0.0 & 0.0 & 6.8 & 0.0 & 0.0 & 0.0 & 0.0 & 0.0 & 1.1 & 0.0 & 64.5 & 0.3 & \textcolor{blue}{\textbf{4.5}} & \textcolor{blue}{\textbf{19.7}} & 2.8 & 4.2 & 9.2 & $\Uparrow$  26.5\\

		ILT \cite{ILT}  + FL & 82.7 & 8.3 & 0.3 & 0.0 & 11.5 & 0.0 & 1.0 & 0.3 & 0.0 & 0.0 & 0.0 & 0.0 & 0.0 & 8.9 & \textcolor{blue}{\textbf{0.5}} & 65.0 & 0.8 & \textcolor{deepred}{\textbf{15.9}} & 12.1 & 4.6 & 4.9 & 10.3 & $\Uparrow$  25.4\\

		MiB \cite{MIB} + FL & 82.2 & 26.1 & \textcolor{blue}{\textbf{32.0}} & 1.7 & 5.8 & 35.3 & 7.8 & \textcolor{blue}{\textbf{72.5}} & 58.4 & 4.1 & 0.0 & 0.0 & 15.8 & 14.4 & \textcolor{deepred}{\textbf{12.3}} & \textcolor{deepred}{\textbf{74.7}} & \textcolor{deepred}{\textbf{2.9}} & 0.0 & 17.5 & 0.0 & 12.7 & 22.7 & $\Uparrow$  13.0\\

		PLOP \cite{PLOP} + FL & 82.6 & \textcolor{blue}{\textbf{76.2}} & \textcolor{deepred}{\textbf{33.9}} & 39.1 & \textcolor{blue}{\textbf{57.3}} & 55.9 & 39.1 & 71.8 & 48.2 & 0.3 & 0.0 & \textcolor{deepred}{\textbf{0.7}} & 7.7 & \textcolor{blue}{\textbf{15.6}} & 0.0 & 61.6 & 0.0 & 0.0 & 13.1 & \textcolor{deepred}{\textbf{9.6}} & 10.5 & 29.7 & $\Uparrow$  6.0\\

		RCIL \cite{RCIL} + FL & 81.1 & 59.2 & 31.9 & \textcolor{blue}{\textbf{43.0}} & \textcolor{deepred}{\textbf{60.3}} & \textcolor{deepred}{\textbf{64.3}} & \textcolor{deepred}{\textbf{63.5}} & \textcolor{deepred}{\textbf{81.5}} & \textcolor{deepred}{\textbf{74.2}} & \textcolor{blue}{\textbf{5.6}} & 0.0 & \textcolor{blue}{\textbf{0.2}} & \textcolor{blue}{\textbf{35.1}} & 4.0 & 0.2 & 66.2 & \textcolor{blue}{\textbf{2.6}} & 0.0 & 9.0 & \textcolor{blue}{\textbf{5.8}} & \textcolor{deepred}{\textbf{19.3}} & \textcolor{blue}{\textbf{33.7}} & $\Uparrow$  2.0\\
  
		\midrule

		\textbf{FBL} (Ours) & \textcolor{deepred}{\textbf{84.2}} & \textcolor{deepred}{\textbf{80.6}} & 28.7 & \textcolor{deepred}{\textbf{64.8}} & 54.2 & \textcolor{blue}{\textbf{62.7}} & \textcolor{blue}{\textbf{58.3}} & 66.6 & \textcolor{blue}{\textbf{72.5}} & \textcolor{deepred}{\textbf{8.4}} & 0.0 & 0.0 & \textcolor{deepred}{\textbf{37.4}} & \textcolor{deepred}{\textbf{22.1}} & 0.0 & \textcolor{blue}{\textbf{69.5}} & 0.0 & 0.0 & \textcolor{deepred}{\textbf{22.3}} & 1.3 & \textcolor{blue}{\textbf{15.3}} & \textcolor{deepred}{\textbf{35.7}} & --\\

		\bottomrule
\end{tabular}}
\label{tab: comparison_voc_8_2}
\vspace{-10pt}
\end{table*}

\begin{table*}[tb]
\centering
\setlength{\tabcolsep}{0.85mm}
\caption{Comparisons of mIoU (\%) on ADE20k dataset \cite{8100027} under the setting of 100-10 with overlapped foregrounds. }
\vspace{-10pt}
\scalebox{0.82}{
\begin{tabular}{c|ccccccccccccccc|>{\columncolor{lightgray}}c|>{\columncolor{lightgray}}c}
	\toprule		
	Class ID & 0-10 & 11-20 & 21-30 & 31-40 & 41-50 & 51-60 & 61-70 & 71-80 & 81-90 & 91-100 & 101-110 & 111-120 & 121-130 & 131-140 & 141-150 & mIoU & Imp. \\
	\midrule
	Finetuning + FL & 0.8 & 0.0 & 0.0 & 0.0 & 0.0 & 0.0 & 0.0 & 0.0 & 0.0 & 0.0 & 0.0 & 0.0 & 0.0 & 0.0 & 10.3 & 0.7 & $\Uparrow$ 27.2 \\
	LWF \cite{LWF} + FL & 0.8 & 0.0 & 0.0 & 0.0 & 0.0 & 0.5 & 1.3 & 0.1 & 0.0 &  0.1 & 0.0 & 0.1 & 0.0 & 1.6 & 9.4 & 0.9 & $\Uparrow$ 27.0 \\
	ILT \cite{ILT}  + FL & 0.8 & 0.0 & 0.0 & 0.3 & 0.0 & 0.5 & 1.9 & 0.2 & 0.1 & 0.3 & 0.0 & 0.1 & 0.0 & 0.7 & 4.3 & 0.6 & $\Uparrow$ 27.3 \\
	MiB \cite{MIB} + FL & 59.7 & 41.3 & 42.4 & 32.4 & 27.9 & 36.4 & 28.7 & 28.9 & 30.1 & 14.7 & \textcolor{blue}{\textbf{2.8}} & \textcolor{blue}{\textbf{5.2}} & 5.8 & \textcolor{deepred}{\textbf{6.0}} & \textcolor{deepred}{\textbf{15.4}} & 25.4  & $\Uparrow$ 2.5 \\
	PLOP \cite{PLOP} + FL & 60.7 & 43.4 & 43.8 & 33.7 & 28.8 & \textcolor{blue}{\textbf{37.0}} & \textcolor{blue}{\textbf{31.1}} & 30.3 & \textcolor{blue}{\textbf{32.1}} & 15.9 & 2.4 & \textcolor{blue}{\textbf{5.2}} & \textcolor{deepred}{\textbf{6.8}} & 2.5 & \textcolor{blue}{\textbf{11.5}} & 26.1  & $\Uparrow$ 1.8 \\
	RCIL \cite{RCIL} + FL &  \textcolor{blue}{\textbf{63.0}} &  \textcolor{blue}{\textbf{46.6}} & \textcolor{blue}{\textbf{47.2}} & \textcolor{blue}{\textbf{35.1}} & \textcolor{blue}{\textbf{31.2}} & 36.0 & 30.8 & \textcolor{blue}{\textbf{32.7}} & 28.1 & \textcolor{blue}{\textbf{16.2}} & 0.3 & \textcolor{deepred}{\textbf{12.3}} & \textcolor{blue}{\textbf{6.3}} & 2.6 & 3.5 & \textcolor{blue}{\textbf{26.4}} & $\Uparrow$ 1.5 \\
	\midrule
	\textbf{FBL} (Ours) &  \textcolor{deepred}{\textbf{67.5}} &  \textcolor{deepred}{\textbf{47.9}} &  \textcolor{deepred}{\textbf{48.9}} &  \textcolor{deepred}{\textbf{38.8}} &  \textcolor{deepred}{\textbf{33.2}} &  \textcolor{deepred}{\textbf{42.6}} &  \textcolor{deepred}{\textbf{35.4}} &  \textcolor{deepred}{\textbf{35.2}} &  \textcolor{deepred}{\textbf{32.8}} &  \textcolor{deepred}{\textbf{17.9}} &  \textcolor{deepred}{\textbf{4.3}} & 3.1 & 3.3 & \textcolor{blue}{\textbf{2.7}} & 0.2 & \textcolor{deepred}{\textbf{27.9}} & -- \\
	\bottomrule
\end{tabular}}
\label{tab: comparison_ade_100_10}
\vspace{-12pt}
\end{table*}

In summary, the major objective of the $l$-th local client $\mathcal{S}_l$ to learn the $t$-th segmentation task $\mathcal{T}_l^t$ is expressed as:
\begin{align}	
	\mathcal{L}_{\mr{obj}} = \mathcal{L}_{\mr{FS}}+\lambda_1\mathcal{L}_{\mr{FR}} + \lambda_2 \mathcal{L}_{\mr{POD}},
	\label{eq: overall_optimization}
\end{align}
where $\lambda_1, \lambda_2$ are trade-off parameters, and $\mathcal{L}_{\mr{POD}}$ denotes the local POD loss proposed in PLOP \cite{PLOP} to perform feature distillation. When $t\geq2$, we set $\lambda_1=0.5$ and $\lambda_2=0.0005$ in Eq.~\eqref{eq: overall_optimization} to train local model $\Theta_{l}^{r, t}$; otherwise, we utilize $\mathcal{L}_{\mr{SE}}$ in Eq.~\eqref{eq: classification_loss} to optimize $\Theta_{l}^{r, t}$. To learn the $t$-th segmentation task $\mathcal{T}_l^t$, local clients belonging to $\mathbf{S}_c$ and $\mathbf{S}_n$ share the same objective function (\emph{i.e.}, Eq.~\eqref{eq: overall_optimization}).

\subsection{Task Transition Monitor}\label{sec: task_transition}
When local clients segment new classes consecutively via Eq.~\eqref{eq: overall_optimization}, global sever $\mathcal{S}_g$ requires to automatically identify when and which local clients collect new classes, and then store the latest old global model $\Theta^{t-1}$ to perform $\mathcal{L}_{\mr{FR}}$ and $\mathcal{L}_{\mr{POD}}$. As a result, the accurate selection of the latest old model $\Theta^{t-1}$ is essential to address inter-client heterogeneous forgetting across different local clients brought by Non-IID class distributions, when new foreground classes arrive. However, considering privacy preservation \cite{zhangdense, zhang2023delving}, we don't have human prior about when to obtain new classes in local clients under the FISS settings. To address this challenge, a naive method is to detect whether the labels of current training data have been observed before. Nevertheless, the Non-IID distributions across local clients make it impossible to identify whether the collected data belongs to old classes seen by other clients or new categories. Thus, inspired by \cite{dong2022federated, fang2022learnable}, we design a task transition monitor to automatically recognize when and which local clients collect new categories. At the $r$-th round, when $\mathcal{S}_l$ receives global model $\Theta^{r, t}$, it evaluates the average entropy $\mathcal{I}_l^{r, t}$ on $\mathcal{T}_l^t$: 
\vspace{-1mm}\begin{equation}\vspace{-1mm}\small
	\mathcal{I}_l^{r, t} = \frac{1}{N_l^t}\sum_{i=1}^{N_l^t}\sum_{j=1}^{HW}\mathcal{H}(P_l^t(\mf{x}_{li}^t, \Theta^{r, t})_j),
	\label{equ:task_transition}
\end{equation}
where $\mathcal{H}(P_l^t(\mf{x}_{li}^t, \Theta^{r, t}))\in\mathbb{R}^{H\times W}$ is the entropy map of $\mathbf{x}_{li}^t$, and $\mathcal{H}(P_l^t(\mf{x}_{li}^t, \Theta^{r, t})_j)$ is entropy scalar of the $j$-th pixel. $\mathcal{H}(\mathbf{p}) = \sum_i\mathbf{p}_i\log\mathbf{p}_i$ is entropy measure function. We consider local clients are collecting new classes, if there is a sudden rise for averaged entropy $\mathcal{I}_l^{r, t}$: $\mathcal{I}_l^{r, t} - \mathcal{I}_l^{r-1, t}\geq \tau$. We then update $t$ via $t\leftarrow t + 1$, and automatically store the latest global model $\Theta^{r-1, t}$ at the $(r\!-\!1)$-th global round as old model $\Theta^{t-1}$ to optimize local model $\Theta_{l}^{r, t}$ via $\mathcal{L}_{\mr{obj}}$ in Eq.~\eqref{eq: overall_optimization}. We set $\tau=0.6$ empirically in this paper. The automatic selection of old model $\Theta^{t-1}$ from global aspect is essential to tackle inter-client heterogeneous forgetting via considering Non-IID distributions across local clients.

\subsection{Optimization Procedure}
At the beginning of each global round in each incremental task, all local clients employ Eq.~\eqref{equ:task_transition} to calculate the average entropy of local data, and then some of local clients are randomly selected by global server $\mathcal{S}_g$ to conduct local training at each round. After these chosen clients utilize task transition monitor to accurately recognize new classes, they automatically store the global model learned at the last global round as the old model $\Theta^{t-1}$ to generate confident pseudo labels for old classes via Eq.~\eqref{eq: pseudo_label}, and optimize local model $\Theta_{l}^{r, t}$ via $\mathcal{L}_{\mr{obj}}$ in Eq.~\eqref{eq: overall_optimization}. 
Finally, the updated local models $\Theta_l^{r, t}$ of selected local clients are aggregated as $\Theta^{r+1, t}$ by $\mathcal{S}_g$ for the next round training. 
The supplementary material provides optimization procedure of our FBL model.

\section{Experiments}\label{sec: experiments}
\subsection{Implementation Details}
We utilize two benchmark datasets: Pascal-VOC 2012 \cite{10.1007/s11263-009-0275-4} and ADE20k \cite{8100027} under various experimental settings to analyze effectiveness of our FBL model. For fair comparisons with baseline ISS methods \cite{LWF, ILT, MIB, PLOP, RCIL} under the FISS settings, we follow them to set exactly the same incremental tasks and class order, while using the identical segmentation backbone (\emph{i.e.}, Deeplab-v3 \cite{deeplab} with ResNet-101 \cite{resnet} pretrained on ImageNet dataset \cite{imagenet}). As claimed in \cite{MIB, PLOP, RCIL}, background pixels in the current task may belong to old classes or new classes from future tasks (\emph{i.e.}, background has some overlap with new foreground classes in the future tasks). In the FISS, we consider more challenging settings by assigning more incremental segmentation tasks with overlapped foregrounds. On Pascal-VOC 2012 \cite{10.1007/s11263-009-0275-4}, 15-1, 4-4, and 8-2 settings with overlapped foregrounds respectively consist in 15 classes followed by 1 classes 5 times ($T=6$), learning 4 classes followed by 4 classes 4 times ($T=5$), and 8 classes followed by 2 classes 6 times ($T=7$). Likewise, on ADE20k \cite{8100027}, 100-10 setting with overlapped foregrounds means 100 classes followed by 10 classes 5 times ($T=6$).



We employ SGD optimizer with initial learning rate as $1.0\times10^{-2}$ to train the first base task and
$1.0\times10^{-3}$ to learn incremental tasks. Considering the limitation of GPU overhead, we set initial local clients as $10$, and add $4$ new local clients for each task. We choose $4$ local clients randomly to perform local training with $6$ epochs for VOC \cite{10.1007/s11263-009-0275-4} and 12 epoches for ADE20k \cite{8100027}. On VOC dataset \cite{10.1007/s11263-009-0275-4}, we randomly select $40\%$ images for each client in each segmentation task under 15-1 setting; otherwise, we randomly sample $50\%$ classes from current label space $\mathcal{Y}^t$, and assign $60\%$ samples from these classes to selected local clients under the 4-4 and 8-2 settings. For the 100-10 setting in ADE20k \cite{8100027}, we randomly choose $70\%$ classes from $\mathcal{Y}^t$, and distribute them to selected clients. Following ISS methods \cite{LWF, ILT, MIB, PLOP, RCIL}, we employ mean Intersection over Union (mIoU) as metric, and evaluate mIoU of all classes after learning the last segmentation task (\emph{i.e.}, $t=T$). This metric evaluates the effectiveness to address heterogeneous forgetting and the ability to segment new classes continually.



\begin{figure*}[tb]
	\centering
	\includegraphics[trim = 14mm 73mm 14mm 73mm, clip, width=495pt, height=120pt]
	{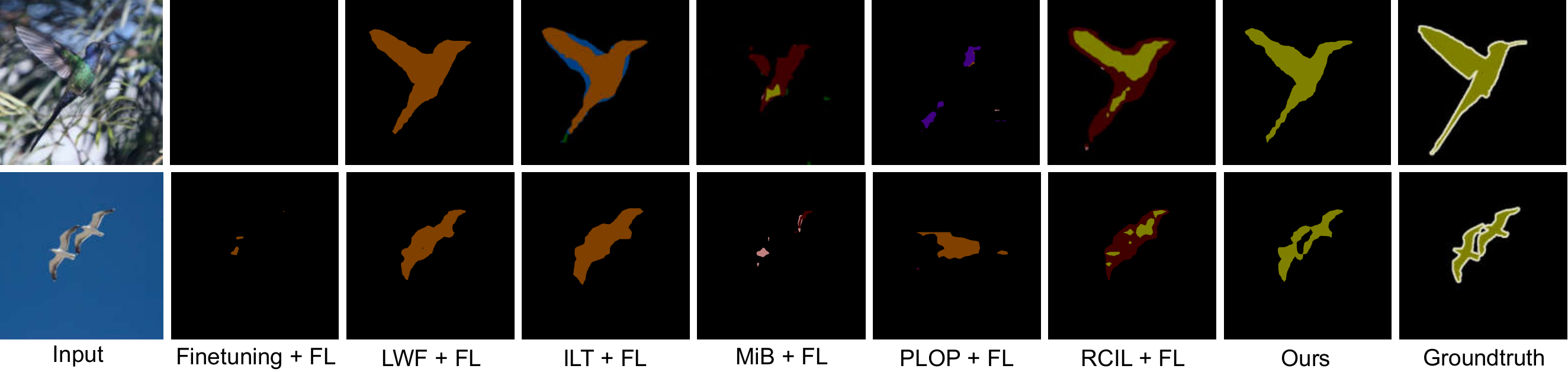}
	\vspace{-20pt}
	\caption{Visualization of some qualitative comparison results on Pascal-VOC 2012 \cite{10.1007/s11263-009-0275-4} under the overlapped 4-4 setting in the FISS. }
	\label{fig: visualization_comparison}
	\vspace{-10pt}
\end{figure*}

\subsection{Comparison Performance}
Experiments on Pascal-VOC 2012 \cite{10.1007/s11263-009-0275-4} and ADE20k \cite{8100027} are introduced to analyze superiority of our model under various settings of FISS, as shown in Tables~\ref{tab: comparison_voc_15_1}$\sim$\ref{tab: comparison_ade_100_10}. Our model achieves large improvements over existing ISS methods \cite{LWF, ILT, MIB, PLOP, RCIL} about $1.5\%\sim51.4\%$ mIoU under various FISS settings. It illustrates the effectiveness of our model against other ISS methods to learn a global continual segmentation model via collaboratively training local models under privacy preservation. Besides, it validates superiority of the proposed loss $\mathcal{L}_{\mr{FS}}$ and $\mathcal{L}_{\mr{FR}}$ to address intra-client and inter-client forgetting heterogeneity in the FISS settings. Some visualization results on Pascal-VOC 2012 \cite{10.1007/s11263-009-0275-4} under the 4-4 setting are shown in Figure~\ref{fig: visualization_comparison}, which verifies the effectiveness of our model to address the FISS problem. 

\begin{figure}[t]
	\centering
	\includegraphics[trim = 39mm 55mm 39mm 55mm, clip, width=235pt, height=100pt]
	{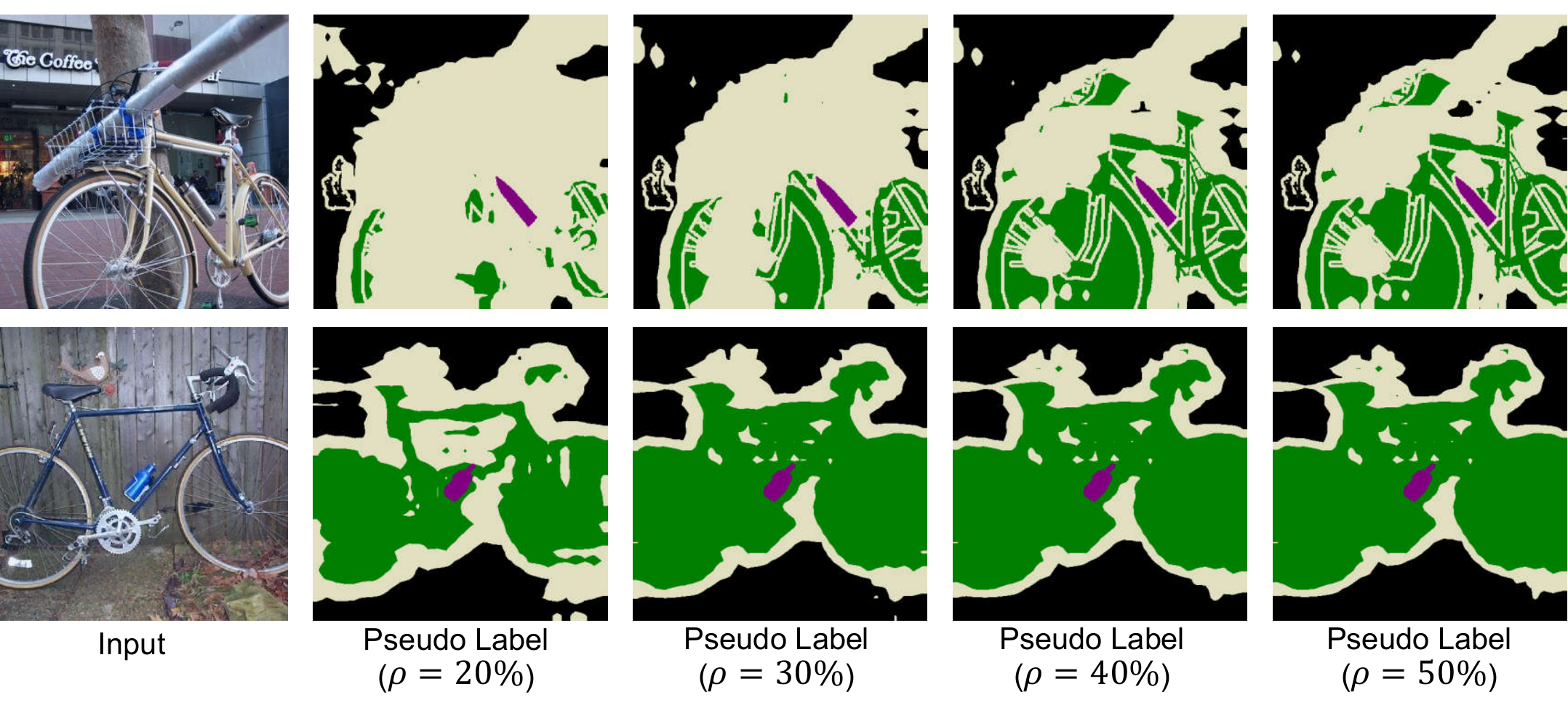}
	\vspace{-25pt}
	\caption{Visualization of some pseudo labels on Pascal-VOC 2012 \cite{10.1007/s11263-009-0275-4} under the 4-4 setting with overlapped foregrounds. }
	\label{fig: pseudo_label}
\end{figure}

\begin{table}[t]
	\centering
	\setlength{\tabcolsep}{1.4mm}
	\caption{Ablation studies on Pascal-VOC 2012 \cite{10.1007/s11263-009-0275-4} under the FISS. }
	\vspace{-10pt}
	\resizebox{\linewidth}{!}{
			\begin{tabular}{c|ccc|cccc|cccc}
				\toprule
				& \multicolumn{3}{c|}{Variants} &\multicolumn{4}{c|}{VOC 4-4 \cite{10.1007/s11263-009-0275-4}} & \multicolumn{4}{c}{VOC 8-2 \cite{10.1007/s11263-009-0275-4}}  \\
				
				Settings & APL& FSC & FRC & 0-16 & 17-20 & mIoU & Imp. & 0-18 & 19-20& mIoU & Imp. \\
				\midrule
				Our-w/oAPL & \xmark & \cmark & \cmark & 41.3 & 34.3 & 40.0 & $\Uparrow$ 3.9& 28.4 & \textcolor{deepred}{\textbf{10.4}} & 26.7 & $\Uparrow$ 9.0\\
				Our-w/oFSC & \cmark & \xmark & \cmark & 41.4 & 33.8 & 40.0 & $\Uparrow$ 3.9 & 31.6 & 6.8 & 29.2 & $\Uparrow$ 6.5 \\
				Our-w/oFRC & \cmark & \cmark & \xmark & 32.6 & 30.8 & 32.3 &  $\Uparrow$11.6& 30.8 & 8.1 & 28.7 & $\Uparrow$ 7.0\\
				
				\rowcolor{lightgray}
				\textbf{FBL} (Ours) &  \cmark & \cmark & \cmark& \textcolor{deepred}{\textbf{45.8}} & \textcolor{deepred}{\textbf{35.8}} & \textcolor{deepred}{\textbf{43.9}} & --& \textcolor{deepred}{\textbf{38.5}} & 8.3 & \textcolor{deepred}{\textbf{35.7}} & --\\
				\bottomrule
		\end{tabular}}
		\label{tab: ablation_studies}
	\end{table}

\begin{table}[t]
\centering
\small
\setlength{\tabcolsep}{2.65mm}
\caption{Task-wise comparisons of mIoU (\%) on Pascal-VOC 2012 dataset \cite{10.1007/s11263-009-0275-4} under the setting of overlapped 4-4 ($T=5$). }
\vspace{-10pt}
\scalebox{0.935}{
	\begin{tabular}{c|ccccc}
		\toprule
		Task ID & t=1 (Base) & t=2 &  t=3 &  t=4 &  t=5   \\
		\midrule
		Finetuning + FL &70.4 & 43.1 & 21.3 & 19.0 & 9.1 \\
		LWF \cite{LWF} + FL &70.4 & 59.8 & 38.7 & 39.1 & 23.8\\
		ILT \cite{ILT}  + FL & 70.4 &56.4 & 36.9 & 35.3 & 22.7 \\
		MiB \cite{MIB} + FL & 70.4 &\textcolor{blue}{\textbf{64.8}} & \textcolor{blue}{\textbf{52.8}} & \textcolor{blue}{\textbf{47.2}} & \textcolor{blue}{\textbf{33.0}} \\
  
		PLOP \cite{PLOP} + FL &70.4& 54.2 & 38.3 & 29.4 & 28.1 \\
		RCIL \cite{RCIL} + FL & 70.5& 60.3 & 40.1 & 36.8 & 32.4 \\
		\midrule
		\rowcolor{lightgray}
		\textbf{FBL} (Ours) & 70.4& \textcolor{deepred}{\textbf{66.6}} & \textcolor{deepred}{\textbf{53.6}} & \textcolor{deepred}{\textbf{49.6}} & \textcolor{deepred}{\textbf{43.9}} \\
		\bottomrule
\end{tabular}}
\label{tab: task_wise_comparison_4_4}
\end{table}

\subsection{Ablation Studies}
To analyze effectiveness of each module in our model, Table~\ref{tab: ablation_studies} presents ablation experiments under various FISS settings. Ours-w/oAPL, Ours-w/oFSC and Ours-w/oFRC indicate the results of our model without adaptive class-balanced pseudo labeling (denoted as APL), forgetting-balanced semantic compensation loss $\mathcal{L}_{\mr{FS}}$ (denoted as FSC) and forgetting-balanced relation consistency loss $\mathcal{L}_{\mr{FR}}$ (denoted as FRC), where Ours-w/oAPL uses constant probability threshold for all old classes to replace adaptive class-specific entropy threshold. When compared with Ours, all ablation variants severely degrade $3.9\%\sim11.6\%$ mIoU. It verifies importance of all modules to address the heterogeneous forgetting. The proposed APL module can effectively tackle background shift via confident pseudo labels, and some confident pseudo labels are visualized in Figure~\ref{fig: pseudo_label}. 

\subsection{Analysis of Task-Wise Comparisons}
As presented in Table~\ref{tab: task_wise_comparison_4_4}, we introduce task-wise comparison results to analyze the effectiveness of our model to address FISS settings. Our model outperforms baseline ISS methods \cite{LWF, ILT, MIB, PLOP, RCIL} for most task-wise comparisons under the overlapped 4-4 setting. The proposed FBL model encourages local clients to learn a global incremental segmentation model cooperatively under privacy preservation. Comparisons in Table~\ref{tab: task_wise_comparison_4_4} show large mIoU improvements of our model to address the FISS problem over other ISS methods. When segmenting new foreground classes consecutively, our model can effectively tackle intra-client and inter-client heterogeneous forgetting on different old classes.

\section{Conclusion}
\label{sec:conclusion}
In this work, we propose a Federated Incremental Semantic Segmentation (FISS) problem, and develop a novel Forgetting-Balanced Learning (FBL) model to address intra-client and inter-client heterogeneous forgetting on old classes. To tackle intra-client heterogeneous forgetting, we design a forgetting-balanced semantic compensation loss and a forgetting-balanced relation consistency loss, under the guidance of adaptive class-balanced pseudo labeling. Meanwhile, we propose a task transition monitor to address inter-client heterogeneous forgetting. It can  automatically recognize new classes and store the latest old global model for distillation. Comparison results demonstrate the superiority of our model to tackle the FISS problem. In the future, we will consider using only few samples of new classes to address intra-client and inter-client forgetting.

{\small
\bibliographystyle{ieee_fullname}
\bibliography{FISS}
}

\newpage

\appendix
\section{Appendix}

\subsection{Optimization Procedure}\label{sec: optimization}
The optimization pipeline of our FBL model to address the FISS problem is presented in \textbf{Algorithm}~\ref{alg: pipline_FBL}. Starting from the first segmentation task, all local clients employ Eq.~\eqref{equ:task_transition} to calculate the average entropy $\mathcal{I}_l^{r, t}$ of local training data $\mathcal{T}_l^t$ at the beginning of each global round, and then some of local clients are randomly selected by global server $\mathcal{S}_g$ to perform local training for each global round. After these selected local clients utilize task transition monitor to accurately recognize new classes, they automatically store the global model learned at the last global round as the old model $\Theta^{t-1}$ to generate confident pseudo labels for old classes via Eq.~\eqref{eq: pseudo_label}, and optimize local model $\Theta_{l}^{r, t}$ via $\mathcal{L}_{\mr{obj}}$ in Eq.~\eqref{eq: overall_optimization} at the $r$-th global round. 
Finally, the updated local models $\Theta_l^{r, t}$ of selected local clients are aggregated as global model $\Theta^{r+1, t}$ by global server $\mathcal{S}_g$, and $\Theta^{r+1, t}$ will be distributed to local clients for the next round training.

\begin{algorithm}[!h]
	\caption{Optimization of The FBL Model.}
	\label{alg: pipline_FBL}
	\LinesNumbered 
	\textbf{Input:} In the $t$-th ($t\geq2$) task, global server $\mathcal{S}_g$ randomly select $w$ local clients  $\{\mathcal{S}_{l_1}, \mathcal{S}_{l_2}, \cdots, \mathcal{S}_{l_w}\}$ with their local datasets as $\{\mathcal{T}_{l_1}^t, \mathcal{T}_{l_2}^t, \cdots, \mathcal{T}_{l_w}^t\}$ at the $r$-th global round; The global server $\mathcal{S}_g$ transmits the latest global model $\Theta^{r, t}$ to selected local clients;
	
	\textcolor{blue}{\textbf{All Local Clients:}} \\
	\For{$\mathcal{S}_l$ in $\{\mathcal{S}_1, \mathcal{S}_2, \cdots, \mathcal{S}_L\}$}{
		Calculate averaged entropy $\mathcal{I}_l^{r, t}$ of local training data $\mathcal{T}_{l}^t$ via Eq.~\eqref{equ:task_transition}; \\
	}
	
	\textcolor{blue}{\textbf{Selected Local Clients:}} \\
	Obtain $\Theta^{r, t}$ from $\mathcal{S}_g$ as the local segmentation model $\Theta_l^{r, t}$;\\
	\For{$\mathcal{S}_{l}$ in $\{\mathcal{S}_{l_1}, \mathcal{S}_{l_2}, \cdots, \mathcal{S}_{l_w}\}$}{
		\emph{Task} = False\;
		\If{$\mathcal{I}_{l}^{r, t} - \mathcal{I}_{l}^{r-1, t} \geq \tau$}{
			\emph{Task} = True\;
		}
		\If{Task = \emph{True}}{
			Store the latest global model $\Theta^{r, t}$ as old model $\Theta^{t-1}$ for local client $\mathcal{S}_l$\;
		}
		\For{$\{\mathbf{x}_{li}^t, \mathbf{y}_{li}^t\}_{i=1}^B$ in $\mathcal{T}_{l}^t$}{
			Generate confident pseduo labels via Eq.~\eqref{eq: pseudo_label}; \\
			Update local model $\Theta_{l}^{r, t}$ via Eq.~\eqref{eq: overall_optimization}; \\
		}
	}
	\textcolor{blue}{\textbf{Global Server:}} \\
	$\mathcal{S}_g$ aggregates the parameters of all local models $\Theta_l^{r, t}$ as $\Theta^{r+1, t}$ for the training of next global round. 
\end{algorithm}

\begin{table*}[t]
\centering
\setlength{\tabcolsep}{1.4mm}
\caption{Ablation studies on Pascal-VOC 2012 dataset \cite{10.1007/s11263-009-0275-4} under the 4-4 and 8-2 settings with overlapped foregrounds. }
\vspace{-10pt}
\resizebox{\linewidth}{!}{
    \begin{tabular}{c|ccc|cccccc|cccccccc}
        \toprule
        & \multicolumn{3}{c|}{Variants} &\multicolumn{6}{c|}{VOC Overlapped 4-4 \cite{10.1007/s11263-009-0275-4} } & \multicolumn{8}{c}{VOC Overlapped 8-2 \cite{10.1007/s11263-009-0275-4} }  \\
        Settings & APL& FSC & FRC & t=1 (Base) & t=2& t=3&t=4&t=5 & Imp. &t=1 (Base) & t=2& t=3&t=4&t=5&t=6& t=7 & Imp.\\
        \midrule
        Our-w/oAPL & \xmark & \cmark & \cmark & 70.4 & \textcolor{deepred}{\textbf{67.6}} & \textcolor{blue}{\textbf{53.3}} & \textcolor{deepred}{\textbf{49.8}} & \textcolor{blue}{\textbf{40.0}}&$\Uparrow$ 3.9& 80.4 & \textcolor{blue}{\textbf{65.1}} & 54.5& 39.5 & \textcolor{blue}{\textbf{41.5}} & 32.9 & 26.7& $\Uparrow$ 9.0 \\

        Our-w/oFSC & \cmark & \xmark & \cmark & 70.4 & \textcolor{blue}{\textbf{67.3}} & 52.2 & 45.8 &\textcolor{blue}{\textbf{40.0}}&$\Uparrow$ 3.9 & 80.4 & \textcolor{deepred}{\textbf{65.8}} & 55.0& 40.8 & 40.6 & 32.5 & \textcolor{blue}{\textbf{29.2}}& $\Uparrow$ 6.5 \\
        Our-w/oFRC & \cmark & \cmark & \xmark & 70.4 & 61.4& 43.2 & 41.4 & 32.3&$\Uparrow$ 11.6 & 80.4 & \textcolor{blue}{\textbf{65.1}} & \textcolor{blue}{\textbf{57.9}}& \textcolor{blue}{\textbf{43.5}} & 41.0 & \textcolor{blue}{\textbf{33.8}} & 28.7& $\Uparrow$ 7.0 \\
        
        \rowcolor{lightgray}
        \textbf{FBL} (Ours) &  \cmark & \cmark & \cmark& 70.4 & 66.6& \textcolor{deepred}{\textbf{53.6}} & \textcolor{blue}{\textbf{49.6}} & \textcolor{deepred}{\textbf{43.9}}&-- & 80.4 & 65.0 & \textcolor{deepred}{\textbf{58.1}}& \textcolor{deepred}{\textbf{47.3}} & \textcolor{deepred}{\textbf{45.8}} & \textcolor{deepred}{\textbf{39.4}} & \textcolor{deepred}{\textbf{35.7}}& -- \\
        \bottomrule
\end{tabular}}
\label{tab: ablation_studies_8_2}
\end{table*}

\subsection{Ablation Studies}\label{sec: ablation_studies}
In this subsection, we present qualitative ablation studies to verify the effectiveness and superiority of our proposed modules. As shown in Table~\ref{tab: ablation_studies_8_2}, when removing one of the designed modules, the performance in terms of mIoU heavily degrades about $3.9\%\sim11.6\%$. Specifically, when compared with Ours, Ours-w/oAPL decreases $3.9\%\sim9.0\%$ mIoU, which validates the effectiveness of the proposed adaptive class-balanced pseudo labeling to mine confident pseudo labels of old classes. These pseudo labels provide strong guidance for two forgetting-balanced losses to address intra-client heterogeneous forgetting on old classes. Moreover, Ours significantly outperforms Ours-w/oFSC by a large margin of $3.9\%\sim6.5\%$ mIoU. This significant performance improvement verifies that our FBL model could effectively tackle forgetting heterogeneity of different old classes within each local client via the forgetting-balanced semantic compensation loss. In addition, Ours-w/oFRC degrades the segmentation performance of $7.0\%\sim11.6\%$ mIoU, compared with Ours. This phenomenon illustrates the effectiveness and superiority of the proposed forgetting-balanced relation consistency loss to compensate heterogeneous relation distillation gains. More importantly, the performance degradation illustrates that all designed modules are effective to collaboratively learn a global incremental segmentation model under the practical FISS settings.

\begin{figure*}[t]
	\centering
	\includegraphics[trim = 14mm 44mm 14mm 44mm, clip, width=495pt, height=220pt]
	{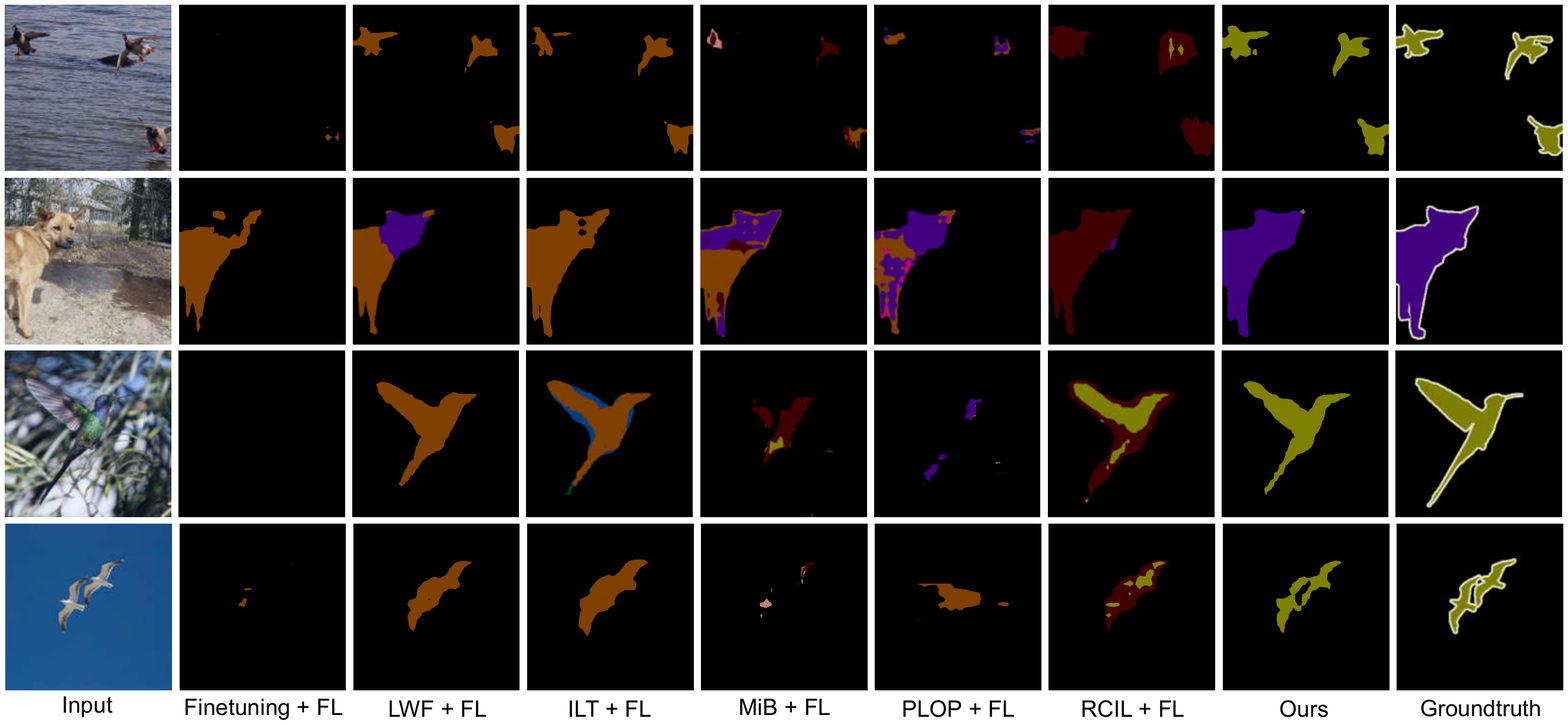}
	\vspace{-20pt}
	\caption{Visualization of some qualitative comparison results on Pascal-VOC 2012 \cite{10.1007/s11263-009-0275-4} under the overlapped 4-4 setting of the FISS. }
	\label{fig: visualization_comparison_dis_4_4}
\end{figure*}

\end{document}